%% file: arxiv.tex
\documentclass[10pt,twocolumn,letterpaper]{article}

\usepackage[pagenumbers]{cvpr}
\input{preamble}

\definecolor{cvprblue}{rgb}{0.21,0.49,0.74}
\usepackage[pagebackref,breaklinks,colorlinks,citecolor=cvprblue]{hyperref}
\usepackage{url}
\usepackage{graphicx}
\usepackage{amsmath}
\usepackage{amssymb}
\usepackage{booktabs}
\usepackage{color}
\usepackage{colortbl}
\usepackage{xcolor}
\usepackage{bbm}
\usepackage{bm}
\usepackage{graphicx}
\usepackage{multirow}
\usepackage{float}
\usepackage{makecell}
\usepackage[title]{appendix}

\title{Customize your NeRF: Adaptive Source Driven 3D Scene Editing via Local-Global Iterative Training}

\author{Runze He$^{1,2}$\footnotemark[1] \quad
Shaofei Huang$^{1,2}$\footnotemark[1] \quad
Xuecheng Nie$^{3}$ \quad
Tianrui Hui$^{1,2}$ \quad \\
Luoqi Liu$^{3}$ \quad
Jiao Dai$^{1,2}$ \quad
Jizhong Han$^{1,2}$ \quad
Guanbin Li$^{4}$\footnotemark[2] \quad
Si Liu$^{5}$\footnotemark[2] \\
$^{1}$Institute of Information Engineering, Chinese Academy of Sciences \quad \\
$^{2}$School of Cyber Security, University of Chinese Academy of Sciences \quad \\
$^{3}$MT Lab, Meitu Inc. \quad
$^{4}$Sun Yat-sen University \quad
$^{5}$Beihang University \quad
\\
{\tt\small \{hrz010109, nowherespyfly, huitianrui\}@gmail.com \quad \{nxc, llq5\}@meitu.com}\\
{\tt\small \{hanjizhong, daijiao\}@iie.ac.cn \quad liguanbin@mail.sysu.edu.cn \quad liusi@buaa.edu.cn}
\\
{\fontsize{10}{10}\selectfont \url{https://customnerf.github.io/}}
}

\begin{document}
\input{sec_arxiv/teaser.tex}
\renewcommand{\thefootnote}{\fnsymbol{footnote}}
\footnotetext[1]{Equal contribution.}
\footnotetext[2]{Corresponding authors.}

\input{sec_arxiv/0_abstract}
\input{sec_arxiv/1_intro}
\input{sec_arxiv/2_related}
\input{sec_arxiv/3_method}
\input{sec_arxiv/4_exp}
\input{sec_arxiv/5_conclusion}

{
    \small
    \bibliographystyle{ieeenat_fullname}
    \bibliography{main}
}

\input{sec_arxiv/appendix}

\end{document}

%% file: preamble.tex
\usepackage[dvipsnames]{xcolor}


%% file: sec_arxiv/teaser.tex
\twocolumn[{
\renewcommand\twocolumn[1][]{#1}
\maketitle
\begin{center}
    \centering
    \vspace*{-.8cm}
    \includegraphics[width=.95\textwidth]{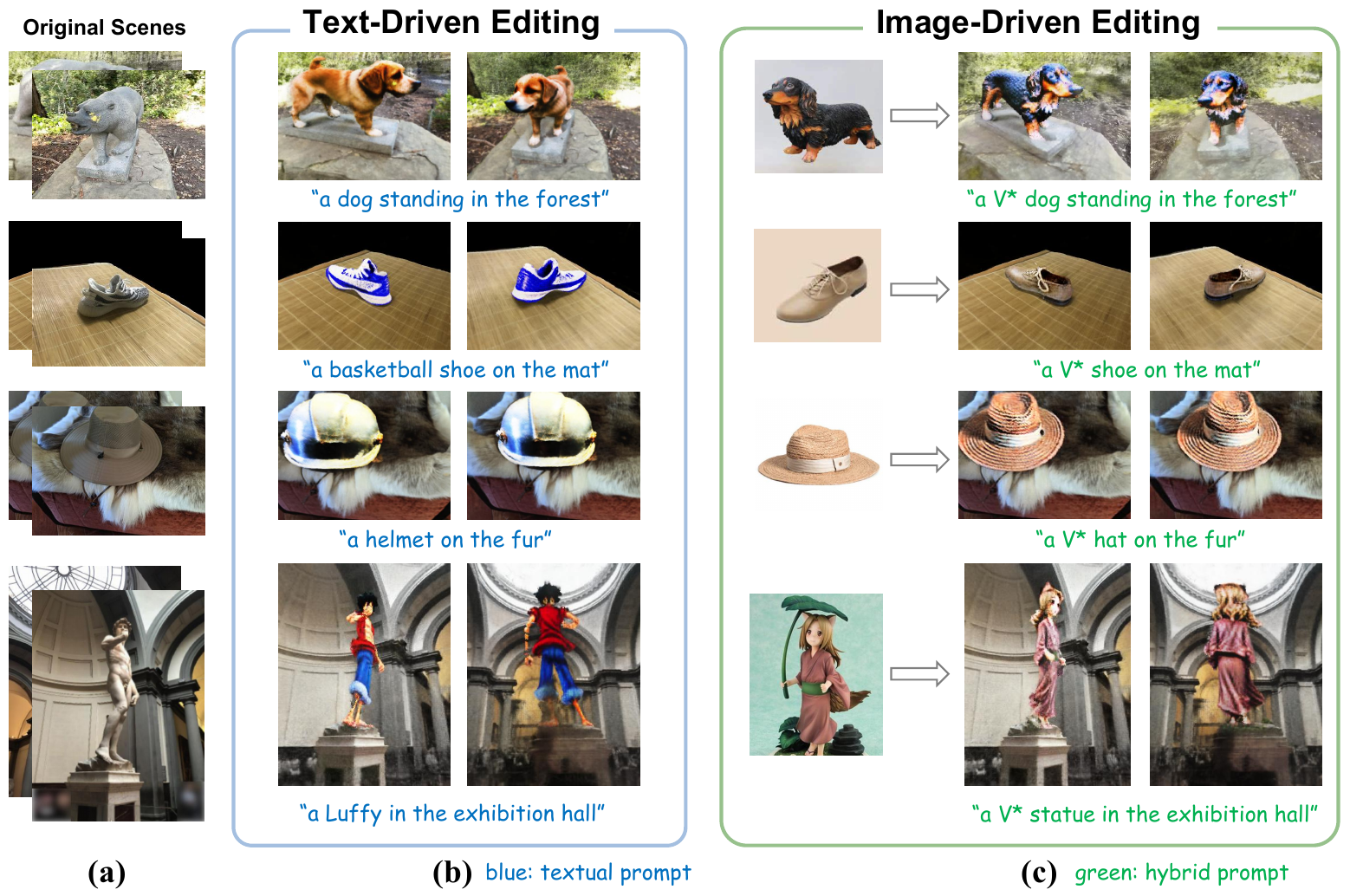}
    \vspace*{-.2cm}
    \captionof{figure}{Adaptive source driven editing by CustomNeRF. Our method can accurately edit the original NeRF scene according to text descriptions or reference images, which are formulated as textual prompts or hybrid prompts. After editing, the foreground regions exhibit appropriate geometric and texture modifications consistent with the editing prompts. The layouts and background contents of the edited scenes remain almost the same as the original ones.}
\label{fig:teaser}
\end{center}
}]

%% file: sec_arxiv/0_abstract.tex
\begin{abstract}
In this paper, we target the adaptive source driven 3D scene editing task by proposing a CustomNeRF model that unifies a text description or a reference image as the editing prompt.
However, obtaining desired editing results conformed with the editing prompt is nontrivial since there exist two significant challenges, including accurate editing of only foreground regions and multi-view consistency given a single-view reference image.
To tackle the first challenge, we propose a Local-Global Iterative Editing (LGIE) training scheme that alternates between foreground region editing and full-image editing, aimed at foreground-only manipulation while preserving the background.
For the second challenge, we also design a class-guided regularization that exploits class priors within the generation model to alleviate the inconsistency problem among different views in image-driven editing.
Extensive experiments show that our CustomNeRF produces precise editing results under various real scenes for both text- and image-driven settings.
\end{abstract}

%% file: sec_arxiv/1_intro.tex
\section{Introduction}
Characterized by the ease of optimization and continuous representation, neural fields (\textit{e.g.}, NeRF~\cite{nerf} and NeuS~\cite{neus}) have garnered extensive attention, further inspiring numerous studies in the field of 3D scene editing, such as retexturing~\cite{texture,xiang2021neutex}, stylization~\cite{zhang2022arf}, or deformation~\cite{garbin2022voltemorph,xu2022deforming} of 3D objects or scenes.
In order to enhance the flexibility and accessibility of editing, recent works~\cite{in2n,dreameditor,voxe,repaintnerf} have proposed text-driven 3D scene editing, utilizing textual prompts to enable editing with more freedom.
These methods take advantage of the powerful capabilities of pre-trained diffusion models~\cite{stablediff,ipix2pix} to provide visual information conformed with the textual prompts, thereby achieving semantic-aware editing.

However, these text-driven methods are only able to achieve \textbf{general editing} to the original 3D scene, leading to a wide range of possible editing outcomes.
To meet more customized requirements of users, a reference image could be introduced to depict the desired outcome precisely, inspiring a more challenging but under-explored task of \textbf{specific editing}.
In Figure~\ref{fig:teaser}, the textual prompt ``\textit{a dog standing in the forest}" is an example of general editing, which replaces the bear statue with an arbitrary dog.
By further restricting the visual characteristics of the dog with a reference image such as a black dog, one can realize specific editing to replace the bear statue with the desired one.

In this paper, we present a unified framework named CustomNeRF to support adaptive source driven 3D scene editing, where a subject-aware Text-to-Image (T2I) model~\cite{CustomDiff} is employed to embed the specific visual subject $V^*$ contained in the reference image into a hybrid prompt, meeting both general and specific editing requirements as illustrated in Figure~\ref{fig:teaser}.
The key to obtaining desired editing results lies in the precise identification of the foreground regions, thereby facilitating view-consistent foreground editing while maintaining the background.
This poses two significant challenges:
Firstly, since NeRF is an implicit representation of 3D scenes, editing NeRF directly is difficult in terms of accurately locating and manipulating only the foreground regions, which necessitates a well-designed training scheme.
Additionally, the lack of accurate inter-view calibration data during training may lead to inconsistency among views, \textit{i.e.}, the Janus problem.
This problem worsens in the image-driven setting, as the diffusion model fine-tuned tends to overfit the input viewpoint of the reference images as encountered in~\cite{dreambooth3d}.

To tackle the first challenge, we propose a Local-Global Iterative Editing (LGIE) training scheme that alternates between foreground region editing and full-image editing.
Concretely, a foreground-aware NeRF is designed to estimate the editing probability in addition to color and density to distinguish between foreground and background regions, enabling the separate rendering of the foreground region.
In the local stage, we feed the foreground-only rendered images and corresponding editing prompt into the T2I model, so that the parameters of NeRF are optimized to only edit the foreground region.
In the global stage, the T2I model takes the fully rendered images and editing prompt as input to optimize NeRF, leveraging useful clues about sizes, positions, and postures contained in the background region to prevent layout drifting from the original scene.
Through the local-global alternating, we can simultaneously constrain the foreground layout and maintain the background content.

For the second challenge, we design a class-guided regularization that only uses class words to denote the subject of the reference image to form a textual prompt during the local editing stage.
By this means, the general class priors within the T2I model can be exploited to promote geometrically consistent editing of the novel subject.
Together with the specific subject appearance learned by the hybrid prompt in the global editing stage, we can gradually transfer the subject's visual characteristics to different views for producing view-consistent editing results.

Our contributions are summarized as follows: (1) We present a unified framework CustomNeRF to support adaptive source driven 3D scene editing, achieving accurate editing in both text-driven and image-driven settings.
(2) A Local-Global Iterative Editing (LGIE) training scheme is proposed to accurately edit foreground regions while maintaining background content.
(3) We design a class-guided regularization to exploit class priors within the T2I model for alleviating the Janus problem in image-driven editing.

%% file: sec_arxiv/2_related.tex
\section{Related Works}

\subsection{Text-to-Image Generation and Editing}
Text-to-image generation~\cite{dalle2,imagen,stablediff,muse,ramesh2021zero,razavi2019generating,brock2018large,karras2019style}, which aims to generate high-quality images according to given text prompts, has gained considerable attention recently.
Among these methods, diffusion model based methods, such as Stable Diffusion~\cite{stablediff} and DALL·E 2~\cite{dalle2}, have shown remarkable image generation ability as a result of training on massive text-image datasets.
Utilizing the rich cross-modal semantic knowledge inherent in pre-trained text-to-image generation models, several works~\cite{sdedit,blenddiff,ipix2pix,p2p} have employed them in text-guided image editing, which aims to modify given images based on text descriptions.
However, these methods are limited to editing styles or contents of images but fail to perform well when faced with more complex modifications, \textit{e.g.}, layout adjustments, subject posture transformations, etc.

To preserve subject identity between the generated and original images while achieving more intricate control over the appearance and posture of these subjects, some works~\cite{dreambooth,TI,CustomDiff} have explored the task of subject-aware text-to-image generation.
In these methods, a small set of images that feature the same subjects are provided for reference and the generation is guided by both text descriptions and the reference images.
DreamBooth~\cite{dreambooth} and TI~\cite{TI} adopt a similar roadway by building upon the pretrained Stable Diffusion model and finetuning it using reference images to map the subject into a unique identifier.
During the testing phase, the learned identifier is injected into the text descriptions for personalized text-to-image generation.
Custom Diffusion~\cite{CustomDiff} realizes faster tuning speed with efficient parameter optimization, and further explores multiple subject combinations through closed-form constrained optimization, thereby extending the capabilities of subject-aware text-to-image generation.
In our work, we employ this method for reference image embedding, thus unifying text-driven and image-driven editing settings.

\subsection{Text-to-3D Generation}
With the significant progress made in text-to-image generation, there has been an increasing exploration of text-to-3D generation where pretrained visual-language models~\cite{clipforge,clipnerf} (\textit{e.g.}, CLIP~\cite{clip}) or text-conditioned diffusion models~\cite{dream3d,dreamfusion,dreambooth3d,SJC} (\textit{e.g.}, Stable Diffusion~\cite{stablediff}) are leveraged to generate realistic 3D objects or scenes from textual descriptions.
DreamField~\cite{dreamfield} utilizes aligned image and text embeddings generated by CLIP to optimize the neural field.
DreamFusion~\cite{dreamfusion} proposes a Score Distillation Sampling (SDS) loss to use a pre-trained 2D diffusion model as a prior for optimization of neural field, which is then improved in Magic3D~\cite{magic3d} with a two-stage optimization framework.
These methods generate 3D objects or scenes from scratch, which cannot be used to edit existing real-world scenes due to the lack of full alignment between images and text descriptions.

\subsection{Neural Field Editing}
Given a pre-trained NeRF of a photo-realistic scene, neural field editing allows for manipulation at varying degrees, resulting in an edited NeRF that is visually coherent and cross-view continuous~\cite{yang2022neumesh,noguchi2021neural,editnerf,in2n,voxe,dreameditor}.
TEXTure~\cite{texture} proposes an iterative painting scheme to enable text-guided editing of 3D shape textures.
EditNeRF~\cite{editnerf} optimizes latent codes to simultaneously modify the appearance and shape of objects.
CLIP-NeRF~\cite{clipnerf} and SINE~\cite{sine} leverage prior models to optimize the geometry and texture of NeRF based on text descriptions or exemplar images.
However, these methods are limited in terms of editing freedom and diversity, restricting their ability to re-create 3D objects/scenes.
Leveraging the powerful text comprehension and generation capabilities of text-guided image generation and editing models, some works~\cite{in2n,voxe,dreameditor,repaintnerf} have enabled more complex textual instructions to freely manipulate 3D scenes at local or global levels.
In order to achieve accurate 3D editing, it is critical to determine precise editing regions while avoiding unnecessary modifications to non-editing regions in a 3D scene, which is relatively challenging for the implicit NeRF representation.
To this end, Vox-E~\cite{voxe} and DreamEditor~\cite{dreameditor} adopt explicit 3D representations (\textit{e.g.}, voxel and mesh) and utilize the cross-attention maps to selectively modify the target regions, RePaint-NeRF~\cite{repaintnerf} constrains background contents with a semantic mask.
Compared with these methods, our method edits directly on NeRF and conducts foreground-only edits with a well-designed Local-Global Iterative Editing training scheme.
Furthermore, our method supports both text-driven and image-driven editing, which allows users to freely customize existing NeRF models according to their requirements.

%% file: sec_arxiv/3_method.tex
\section{Preliminary}
\subsection{Latent Diffusion Model}
\label{sec:ldm}
Latent Diffusion Models (LDMs)~\cite{stablediff} is a variation of Denoising Diffusion Probabilistic Models (DDPMs)~\cite{ddpm} that operate in the latent space of images.
It consists of an auto-encoder and a UNet-structured diffusion model.
The auto-encoder is composed of an encoder $\mathcal{E}(\cdot)$ that maps an image $\mathbf{x}$ into a spatial latent code $\mathbf{z}=\mathcal{E}(\mathbf{x})$ and a decoder $\mathcal{D}(\cdot)$ that maps the latent code back to image $\mathcal{D}(\mathcal{E}(\mathbf{x}))$.
The diffusion model is trained to produce latent codes within the encoder's latent space with:
\begin{equation}
\label{eq:ldm}
\mathcal{L}_{LDM}=\mathbb{E}_{\mathbf{z}\sim\mathcal{E}(\mathbf{x}),y,\epsilon\sim\mathcal{N}(0,1),t}\lbrack\lVert\epsilon_\phi(\mathbf{z}_t;t,y)-\epsilon \rVert_2^2\rbrack,
\end{equation}
where $\mathbf{z}_t$ is the noisy latent code at timestep $t$, $\epsilon$ is the sampled noise, $\epsilon_\phi(\cdot)$ is the UNet that predicts the noise content, and $y$ denotes the text description.
During inference, a random noise $\mathbf{z}_T$ is sampled and iteratively denoised to produce $\mathbf{z}_0$, which is mapped back to the image space through image decoder $\mathbf{x}'=\mathcal{D}(\mathbf{z}_0)$.

\begin{figure*}[!t]
    \centering
    \includegraphics[width=\linewidth]{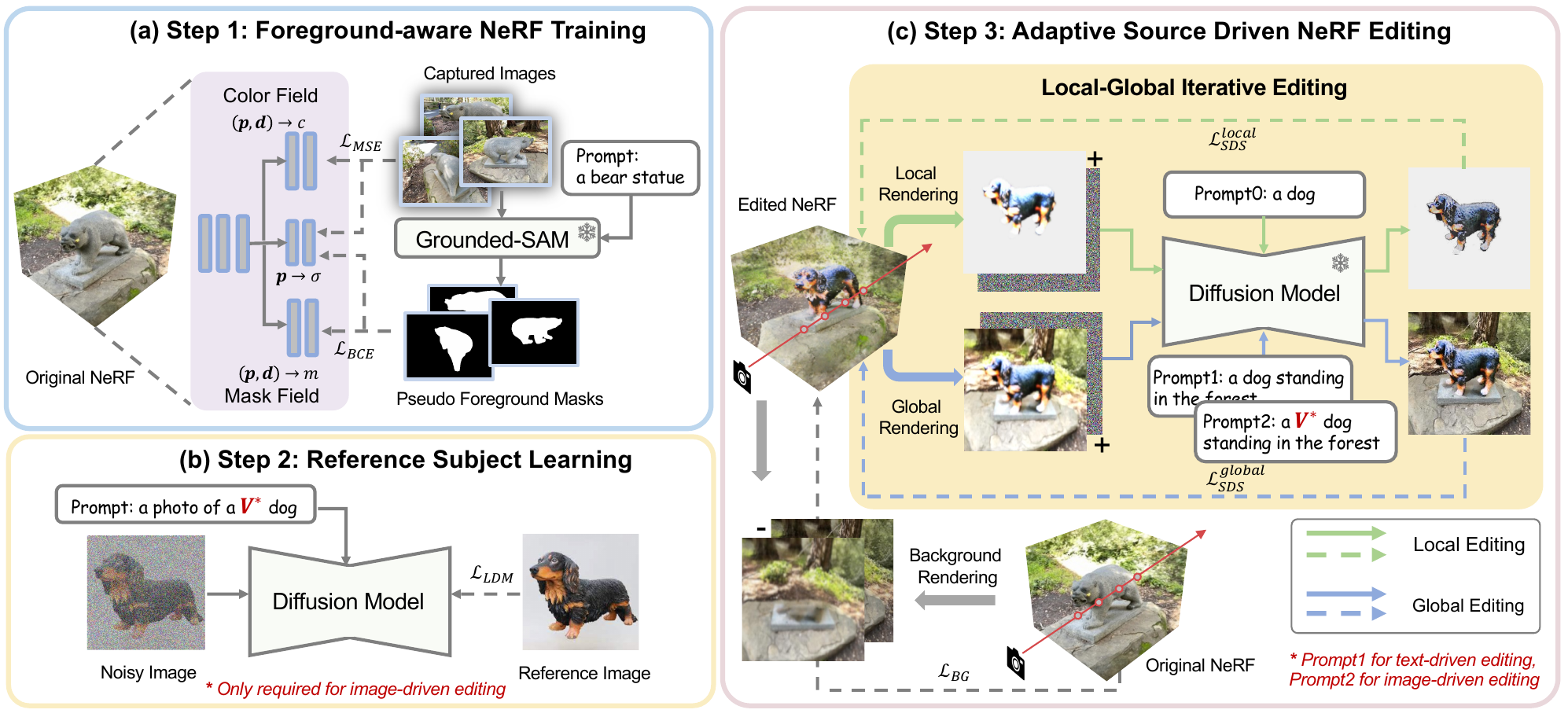}
    \vspace{-1mm}
    \caption{The overall pipeline of our CustomNeRF. Given an original 3D scene, CustomNeRF aims to edit it according to a textual description or a reference image. Our pipeline consists of three steps in total. (a) Reconstruct the original 3D scene with a foreground-aware NeRF, which enables the identification of foreground regions. (b) Finetune the pre-trained text-to-image diffusion model to embed the visual subject in the reference image into a special token $V^*$. This step is only required in image-driven editing. (c) Optimize the given NeRF with our proposed Local-Global Iterative Editing (LGIE) training scheme to align with the textual description or the reference image. The local and global data flows occur in an alternating manner. \textit{Prompt2} is employed in image-driven editing whereas \textit{Prompt1} is utilized in text-driven editing. \textit{Prompt0} is shared in both editing settings.}
    \label{fig:pipeline}
\end{figure*}

\subsection{Score Distillation Sampling Loss}
\label{sec:sds}
Score Distillation Sampling (SDS) loss~\cite{dreamfusion} optimizes a NeRF model under the guidance of a 2D text-to-image diffusion model mentioned in Section~\ref{sec:ldm} to realize text-to-3D generation.
Let $\theta$ denote the parameters of NeRF model, and $g(\cdot)$ denotes the process of image rendering.
The NeRF model first renders a randomly viewed image $\mathbf{x}=g(\theta)$, which is then transformed to noisy latent code $\mathbf{z}_t$ by feeding to the latent encoder $\mathcal{E}(\cdot)$ and adding noise sequentially.
The SDS loss is calculated as follows:
\begin{equation}
\label{eq:sds}
\nabla_\theta \mathcal{L}_{SDS}(\phi,g(\theta))=\mathbb{E}_{t,\epsilon}\lbrack\omega(t)(\epsilon_\phi(\mathbf{z}_t;t,y)-\epsilon)\frac{\partial \mathbf{x}}{\partial \theta}\rbrack,
\end{equation}
where $\omega(t)$ is a weighting function depending on timestep $t$ (Equation~\ref{eq:ldm}).
During training, the above gradients are back-propagated to the NeRF model only, enforcing it to render images that look like outputs of the frozen diffusion model.

\section{Method}
\label{sec:method}
Given a set of well-calibrated images representing a 3D scene, our goal is to edit the reconstructed 3D scene under the guidance of a textual prompt or a reference image, enabling editing of the foreground regions while preserving the background regions simultaneously.
The overall pipeline of our CustomNeRF is illustrated in Figure~\ref{fig:pipeline}, which consists of three steps.
First, to reconstruct the original 3D scene, we train a foreground-aware NeRF where an extra mask field is introduced to estimate the editing probability besides regular color and density prediction as shown in Figure~\ref{fig:pipeline}(a).
The editing probability is utilized in the last step to discriminate regions to be edited (\textit{i.e.}, foreground regions) from irrelevant regions (\textit{i.e.}, background regions).
Second, as illustrated in Figure~\ref{fig:pipeline}(b), we employ a subject-aware Text-to-Image (T2I) diffusion model to encode the subject in the reference image as a special token $V^*$, thus forming a hybrid prompt for image-driven editing (not required for text-driven editing).
Finally, we train the edited NeRF conditioned on the textual prompt or the hybrid prompt using the Local-Global Iterative Editing (LGIE) scheme, allowing it to render images that conform with the editing prompt while preserving background contents, which is presented in Figure~\ref{fig:pipeline}(c). 
The class-guided regularization is further adopted in image-driven editing to alleviate the Janus problem.

\subsection{Foreground-aware NeRF Training}
\label{sec:step1}
Given a 3D point $\bm{p}$ and a view direction $\bm{d}$, a typical NeRF~\cite{mildenhall2021nerf} defines MLPs to output view-dependent color $c(\bm{p},\bm{d})$ and the point density $\sigma(\bm{p})$.
Considering a pixel's camera ray $\bm{r}(k)=\bm{o}+k\bm{d}$ starting from origin $\bm{o}$ in direction $\bm{d}$, with $k$ representing the distance from $\bm{o}$.
The color of this pixel can be obtained by volume rendering:
\begin{equation}
\begin{split}
\label{eq:volume render}
    C(\boldsymbol{r})=\int_{k_{n}}^{k_{f}} T(k) \sigma(\boldsymbol{r}(k)) c(\boldsymbol{r}(k), \boldsymbol{d}) \mathrm{d}k,\\
    \mathrm{where} \quad T(k)=\exp \left(-\int_{k_{n}}^{k} \sigma(\boldsymbol{r}(k)) \mathrm{d}k \right),
\end{split}
\end{equation}

where $k_n$ and $k_f$ are the near plane and the far plane, $T(k)$ can be regarded as transparency. 
MSE loss between the rendered color $\hat{C}(\bm{r})$ and the captured color $C(\bm{r})$ is calculated for supervision.

In our method, to identify foreground regions, we extend the typical NeRF with an extra mask field estimating the editing probability $m(\bm{p}, \bm{d}) \in (0,1)$ of point $\bm{p}$ from the view $\bm{d}$, which is shown in Figure~\ref{fig:pipeline}(a).
Pixel editing probability of ray $\bm{r}$, denoted by $\hat{M}(\bm{r})$ can be rendered in the same way as Equation~\ref{eq:volume render}.
To acquire the ground truth of $\hat{M}(\bm{r})$, we incorporate Grounded SAM~\cite{groundedsam,sam,liu2023grounding} to obtain the binary segmentation mask $M(\bm{r})$ of each captured image according to a textual prompt describing the edited region, \textit{e.g.}, \textit{``a bear statue''} in Figure~\ref{fig:pipeline}(a).
BCE loss is calculated between $\hat{M}(\bm{r})$ and the pseudo label $M(\bm{r})$.

\subsection{Reference Subject Learning}
\label{sec:step2}
In order to unify the formulation of image-driven and test-driven editing tasks, we adopt an efficient subject-aware text-to-image generation model Custom Diffusion~\cite{CustomDiff} for reference subject encoding.
Custom Diffusion is built upon a pre-trained text-conditioned diffusion model and finetuned on several reference images to generate images of the specific subject while preserving its key identifying characteristics.
As illustrated in Figure~\ref{fig:pipeline}(b), a special word $V^*$ is defined to uniquely identify the reference subject in the reference image, which is optimized by the LDM loss (Equation~\ref{eq:ldm}) to map it to a proper word embedding.
After training, $V^*$ can be treated as a regular word token to form a hybrid prompt, \textit{e.g.}, \textit{``a photo of a $V^*$ dog"}, for customized generation of the specific black dog.
In this way, our CustomNeRF is able to conduct consistent and effective editing with adaptive types of source data including images or texts.
More details about the optimization of Custom Diffusion can be found in~\cite{CustomDiff}.

\subsection{Adaptive Source Driven NeRF Editing}
\label{sec:step3}
In this step, we aim to obtain an edited NeRF that can re-create the 3D scene aligning with the text description or the reference image.
An intuitive way to obtain this edited NeRF is to first initialize it as a copy of the original NeRF obtained in Section~\ref{sec:step1} and then optimize it with the SDS loss calculated using the pre-trained T2I model to align with the textual or hybrid prompt.
However, due to the implicit NeRF representation, naively applying SDS loss to the whole 3D scene results in changes to the background regions, which should be kept the same as the original scene.
To this end, we propose a Local-Global Iterative Editing (LGIE) scheme that alternates between foreground region editing and full-image editing.
Furthermore, to mitigate the Janus problem in image-driven editing, we design a class-guided regularization to exploit general class priors contained in the T2I model to inject novel concepts into scenes in a more geometrically consistent way.
Details of the above training scheme can be found in Figure~\ref{fig:pipeline}(c).

\noindent \textbf{Local-Global Iterative Editing.}
Our LGIE training scheme alternates between the local editing stage which separately renders and edits the foreground regions only, and the global editing stage which edits foreground regions contextualized on background clues.
We first take text-driven editing as an example for illustration.
As illustrated in Figure~\ref{fig:pipeline}(c), given a 3D scene of the bear statue, our goal is to edit it with the textual prompt ``a dog standing in the forest'' to replace the bear with a dog.
In the local editing stage, we modify Equation~\ref{eq:volume render} to render foreground regions using the editing probability $m(\bm{r}(k),\bm{d})$:
\begin{equation}
\begin{split}
\label{eq:foreground render}
    C_f(\bm{r})=\int_{k_{n}}^{k_{f}} T_f(k) \sigma_f(\bm{r}(k),\bm{d}) c(\boldsymbol{r}(k), \bm{d}) \mathrm{d}k,\\
    \mathrm{where} \quad T_f(k)=\exp \left(-\int_{k_{n}}^{k} \sigma_f(\bm{r}(k),\bm{d}) \mathrm{d} k\right),
\end{split}
\end{equation}
where point density $\sigma(\bm{r}_t)$ is adjusted by $m(\bm{r}_t,\bm{d})$ to eliminate the background contents in the rendered image as follows:
\begin{equation}
    \sigma_f(\bm{r}(k),\bm{d})=m(\bm{r}(k),\bm{d}))\times \sigma(\bm{r}(k)).
\end{equation}
The rendered foreground image $\mathbf{x}_f$ is then fed into the frozen T2I model together with the foreground part of the above editing prompt (\textit{e.g., ``a dog"}).
Following~\cite{dreamfusion}, view-dependent words, \textit{e.g.}, ``front view", ``side view", and ``back view", \textit{etc}, are appended to the prompt to alleviate Janus problem in text-driven editing.
Due to the absence of a standardized definition of direction in the captured dataset, instead of setting it manually, we conduct matching between the CLIP features of images and those of texts to assign an appropriate view-dependent word to the rendered image.
Afterward, the local SDS loss $\mathcal{L}_{SDS}^{local}$ formulated in Equation~\ref{eq:sds} is calculated to optimize the NeRF parameters. 

In the global editing stage, we render the full image and feed it into the T2I model together with the complete editing prompt ``a dog standing in the forest'' to calculate a global SDS loss
$\mathcal{L}_{SDS}^{global}$.
In this stage, the context information in the background regions serves as a useful clue for editing, realizing foreground-background harmony.
To further avoid changes to the context of the background region, we detach the gradients of the background region in fully rendered images to prevent them from affecting the NeRF parameters.

\noindent \textbf{Class-guided regularization}
For image-driven editing where hybrid prompt (\textit{e.g.}, \textit{``a $V^*$ dog standing in the forest''} is provided to replace the bear statue with a specific black dog, we apply a class-guided regularization method to mitigate the Janus problem which becomes severe due to the single-view reference image.
Concretely, in the local editing stage, we remove the special token $V^*$ from the foreground part of the editing prompt, so that the class word \textit{``dog''} plays a major role in the local optimization process.
In this manner, the general class priors embedded within the T2I model are effectively leveraged to guide geometrically reasonable optimization.
The global stage runs in the same way as the text-driven setting, except that the hybrid prompt is adopted to inform the NeRF model with the appearance information of the reference subject.
By alternating between local and global stages, the appearance of the reference subject can be transferred to various views gradually, producing view-consistent editing results.

\noindent \textbf{Loss functions.}
Apart from the above two SDS losses, we further explicitly constrain the rendered pixel color of the background region to be the same as the original one for the goal of background preservation.
We render the background regions before and after editing, which are denoted as $\mathbf{x}_b^o$ and $\mathbf{x}_b^e$. 
Concretely, we first calculate the background point density $\sigma_b(\bm{r}(k),\bm{d})$ for background rendering:
\begin{equation}
\label{eq:bg render}
    \sigma_b(\bm{r}(k),\bm{d})=(1 - (m(\bm{r}(k),\bm{d})))\times \sigma(\bm{r}(k),
\end{equation}
where subscript $o$ and $e$ are omitted for simplicity.
The background image is rendered using Equation~\ref{eq:foreground render} except that $\sigma_f(\bm{r}(k),\bm{d})$ is replaced with $\sigma_b(\bm{r}(k),\bm{d})$.
An MSE loss $\mathcal{L}_{bg}$ is calculated between $\mathbf{x}_f^o$ and $\mathbf{x}_f^e$ to enforce the edited NeRF rendering similar background contents with the original NeRF.

The overall loss function for the third step is summarized as follows:
\begin{equation}
     \mathcal{L} = \lambda_{SDS}\mathcal{L}_{SDS} + \lambda_{bg}\mathcal{L}_{bg},
\label{eq:loss}
\end{equation}
where $\lambda_{SDS}$ and $\lambda_{bg}$ are hyperparameters to balance different losses, which are empirically set as 0.01 and 1000 respectively.
We use $\mathcal{L}_{SDS}$ to represent both $\mathcal{L}_{SDS}^{local}$ and $\mathcal{L}_{SDS}^{global}$ which are alternately calculated during training.

%% file: sec_arxiv/4_exp.tex
\begin{figure*}[!htbp]
    \centering
    \includegraphics[width=.95\linewidth]{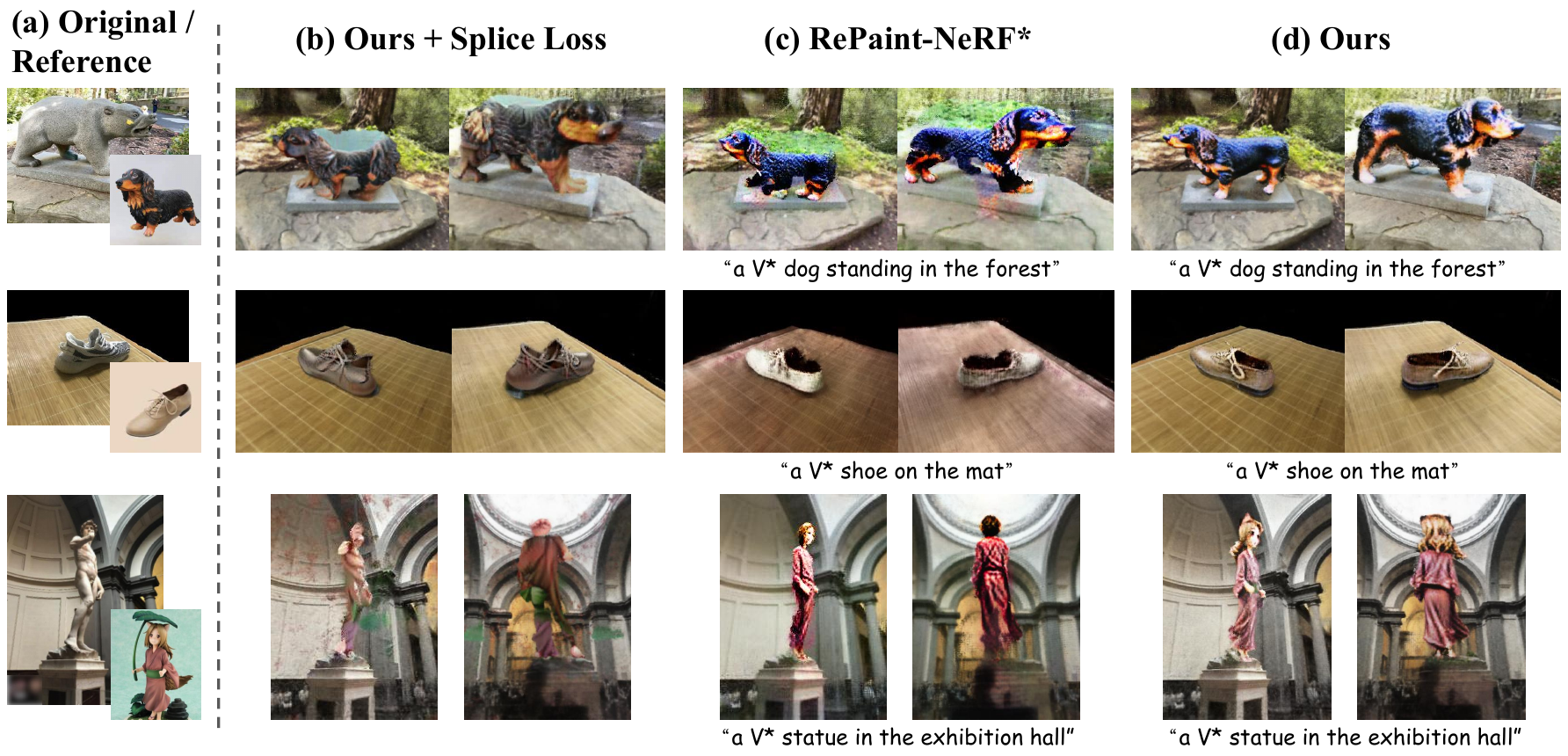}
    \caption{Qualitative comparison on the image-driven editing setting. * denotes replacing the employed T2I model with Custom Diffusion.}
    \label{fig:comp_img}
\end{figure*}

\begin{figure*}[!htbp]
    \centering
    \includegraphics[width=.95\linewidth]{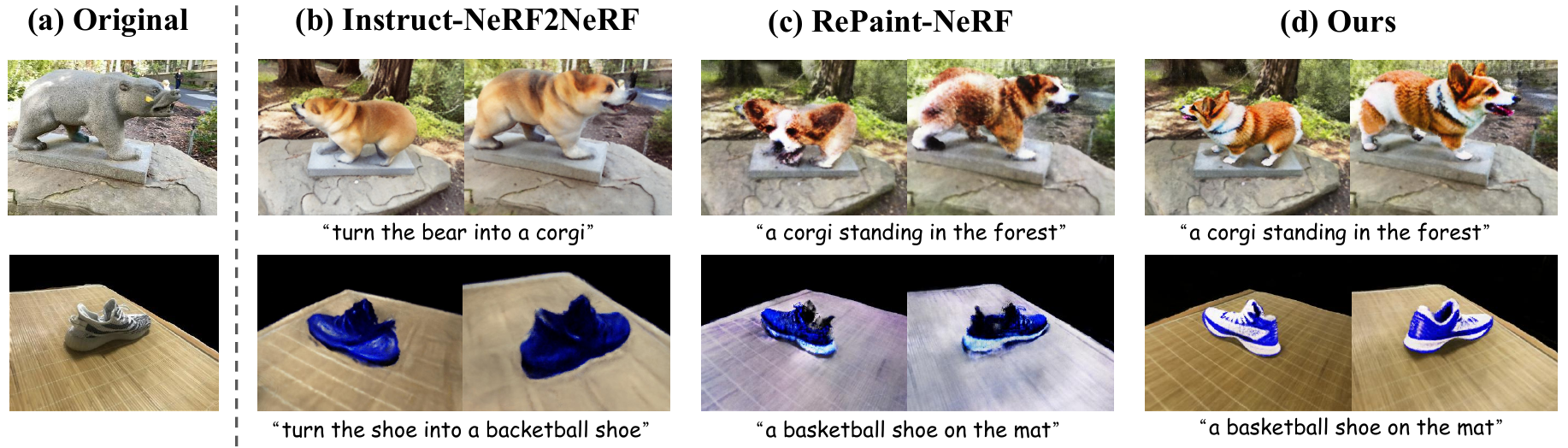}
    \caption{Qualitative comparison on the text-driven editing setting.}
    \label{fig:comp_text}
\end{figure*}

\section{Experiments}
\label{sec:Experiments}

\subsection{Experimental Setup}
\label{sec:exp}
\noindent \textbf{Datasets.}
To validate the effectiveness of our CustomNeRF, we test on 8 real scenes from BlendedMVS~\cite{yao2020blendedmvs}, LLFF~\cite{mildenhall2019local}, IBRNet~\cite{wang2021ibrnet}, and Bear Statue~\cite{in2n}.
These scenes feature complex backgrounds, including outdoor statues and objects frequently encountered in everyday life, \textit{etc}.
For text-driven editing, 4-6 textual prompts are adopted for each scene.
For image-driven editing, we use images downloaded from the Internet as reference images, with 3-5 reference images available for each scene.

\def\in2n{Instruct-NeRF2NeRF}
\def\ip2p{Instruct-pixel2pixel}
\def\rep{RePaint-NeRF}
\def\reps{RePaint-NeRF*}

\noindent \textbf{Baselines.}
For text-driven editing, we compare our method with two recent editing methods based on NeRF representation, including: (1) \in2n{}~\cite{in2n}, which edits NeRF by iteratively updating dataset with images generated by a 2D editing model~\cite{ipix2pix}.
(2) \rep{}~\cite{repaintnerf}, which uses SDS loss for NeRF optimization and keeps non-editing regions unchanged by 2D semantic masks.

For image-driven editing, due to the absence of existing methods for this setting, we modified several existing works to serve as baselines for comparison with our method, including:
(1) Ours+Splice Loss: Splice loss~\cite{tumanyan2022splicing} is proposed to disentangle structure and appearance information from an image for image editing, which is further demonstrated effective to transfer the texture from an exemplar image for NeRF editing in SINE~\cite{sine}.
We apply this loss to our foreground-aware NeRF during the editing stage to transfer appearance information from the reference image while maintaining the structure layout of the original scene.
(2) \reps{}: Since the original framework of \rep{} only supports text-driven editing, we replace its employed T2I model with the Custom Diffusion finetuned on reference images. In this way, the reference image can be embedded as a special token in the editing prompt, thereby equipping \rep{} with the capability for image-driven editing. 

\noindent \textbf{Evaluation metrics.}
Following ~\cite{in2n,dreameditor}, we use the CLIP Text-Image Directional Similarity to evaluate the alignment between the images and text prompts before and after editing, reflecting whether the image editing direction is consistent with the text changing direction. 
The DINO~\cite{oquab2023dinov2} Similarity used in subject-driven image generation is also adopted as a metric.
By calculating the average of the DINO similarity between the edited images from different views and the reference image, we can measure whether the edited NeRF is consistent with the reference image.
Furthermore, to compensate for the lack of objective metrics in subjective editing tasks, we also conduct user studies in text-driven and image-driven editing.

\def\voteImage{75.6}
\def\voteText{85.2}
\def\in2n{Instruct-NeRF2NeRF}
\def\de{DreamEditor}
\def\nerf{NeRF}

\noindent \textbf{Implementation details.}
In our experiments, we employ Instant-NGP~\cite{muller2022instant} to learn the original neural field and make necessary extensions to enable the prediction of editing probabilities.
The training of original NeRF takes 3K iterations with the supervision of both rendered image and editing probability. 
We use Stable Diffusion v1.5 and follow the Custom Diffusion training process to conduct reference subject learning with 250 steps.
In the editing step, we use the Adam optimizer~\cite{kingma2014adam} with a learning rate of  5e-4 to optimize the NeRF parameters. 
The duration of the editing process varies from 10 minutes to 1 hour depending on the complexity of the scene and the editing type on a single NVIDIA RTX 3090 GPU with at most 10K iterations.
We implement our model using PyTorch~\cite{pytorch}.

\subsection{Qualitative Results}
\label{sec:qualitative}
\noindent \textbf{Image-driven editing}.
We provide qualitative results of image-driven editing in Figure~\ref{fig:comp_img} to compare our CustomNeRF with other baseline methods.
It is shown in Figure~\ref{fig:comp_img}(b) that Splice Loss can only transfer superficial attributes such as color and texture from the reference image to the edited NeRF, while it fails to modify the geometry of the edited object. For instance, in the 3rd row, where the statue of David is replaced with a comic character, it only turns the statue into red. 
Although RePaint-NeRF* (Figure~\ref{fig:comp_img}(c)) learns the texture and geometry information of the reference image to some extent, it is still limited in the richness of details and similarity to the reference image, \textit{e.g.}, the hairstyle and clothes texture of the comic character are not presented in the edited NeRF. Moreover, background contents are not well-preserved in this method.
In contrast, our method is able to generate edited results that not only maintain the background effectively but also bear a closer resemblance to the reference image compared to the two baselines.
For example, in the 1st row, our method successfully maintains the ring of yellow fur on the leg of the black dog. In the more challenging scenario depicted in the 3rd row, our approach preserves intricate details such as the hairstyle of the comic character and the folds in the clothes, which further demonstrates the superiority of our CustomNeRF.
More visualization results are included in supplementary materials.

\noindent \textbf{Text-driven editing}.
We also provide qualitative results of text-driven editing in Figure~\ref{fig:comp_text}. 
Compared to baseline methods, our approach is capable of realizing editing with more substantial shape and semantic changes and maintaining consistency between different views in complex scenes, such as transforming a bear statue into a corgi dog. 
Instruct-NeRF2NeRF still retains the bear's body shape and RePaint-NeRF tends to produce abnormal renderings. 
In terms of background preservation, our method also significantly outperforms the baseline approaches, particularly RePaint-NeRF, which tends to have its background styles easily influenced by the foreground content.

\begin{table}[!htbp]
    \begin{center}
  \resizebox{1.0\linewidth}{!}{
  \begin{tabular}{lccc}
    \toprule
    Method  & CLIP$_{dir}$ $\uparrow $  & DINO$_{sim}$ $\uparrow $ & Vote Percentage $\uparrow $\\
    \midrule
    Ours+Splicing Loss & 15.47  & 46.51 & 10.6\%\\
    RePaint-NeRF* & 16.46  & 38.74 & 13.8\%\\
    \textbf{CustomNeRF (Ours)} & \textbf{20.07} & \textbf{47.44} & \textbf{\voteImage{}}\%\\
    \bottomrule
  \end{tabular}}
  \caption{Quantitative comparison on the image-driven editing setting. CLIP$_{dir}$ is short for CLIP Text-Image Directional Similarity and DINO$_{sim}$ is short for DINO Similarity.}
  \vspace{-3mm}
  \label{tab:quantitative-image}
  \end{center}
\end{table}

\begin{table}[!htbp]
\begin{center}
  \resizebox{.9\linewidth}{!}{
  \begin{tabular}{lccc}
    \toprule
    Method  & \makecell{CLIP$_{dir}$ $\uparrow $}  & \makecell{Vote Percentage $\uparrow $ } \\
    \midrule
    Instruct-NeRF2NeRF & 11.48 & 10.6\% \\
    RePaint-NeRF & 16.34 & 4.2\% \\
    \textbf{CustomNeRF (Ours)} & \textbf{22.66} & \textbf{\voteText{}}\% \\
    \bottomrule
  \end{tabular}}
  \caption{Quantitative comparison on the text-driven editing setting. CLIP$_{dir}$ is short for CLIP Text-Image Directional Similarity.}
  \vspace{-6mm}
  \label{tab:quantitative-text}
\end{center}
\end{table}

\subsection{Quantitative results}
\label{sec:quantitative}
We present the quantitative comparison with other methods on the image-driven editing and text-driven editing tasks in Table~\ref{tab:quantitative-image} and Table~\ref{tab:quantitative-text} respectively.
For user study, we distribute 50 questionnaires for both image-driven editing and text-driven editing, including 10 image-driven edits across 5 scenes and 10 text-driven edits across 2 scenes, asking users to choose the best editing results.
Our CustomNeRF significantly outperforms baseline methods and achieves the highest editing scores in both tasks, especially in user evaluation performances.
Although Ours+Splicing Loss in Table~\ref{tab:quantitative-image} also achieves a relatively high DINO Similarity, this is largely due to the fact that it is optimized based on DINO-ViT features, giving it a significant advantage when computing DINO feature similarity. However, as indicated by the other metrics and qualitative results in Figure~\ref{fig:comp_img}, it is substantially inferior in performance compared to our method.

\subsection{Ablation study}
\label{sec:ablation}

\begin{figure}[!t]
    \centering
    \includegraphics[width=\linewidth]{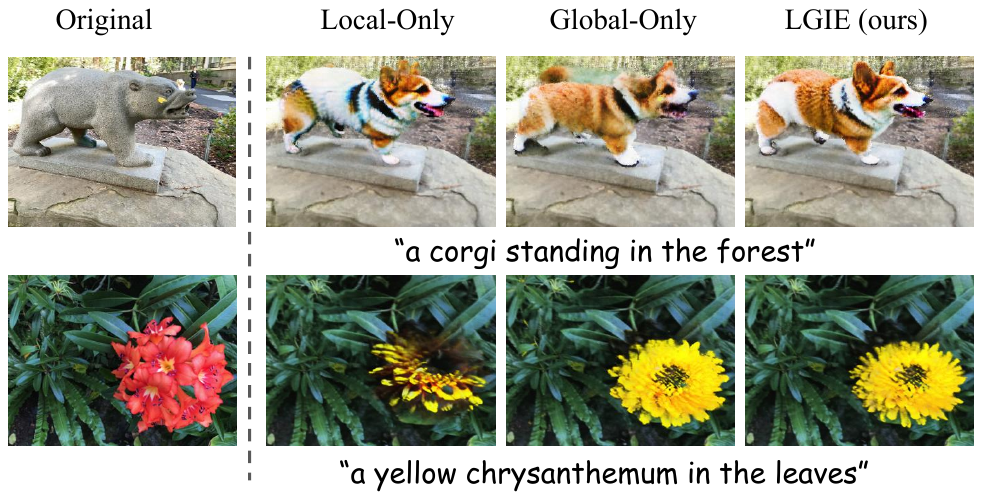}
    \caption{Ablation study of Local-Global Iterative Editing. Local-only and Global-only denote editing with only local SDS loss and global SDS loss respectively.}
    \label{fig:local_global}
\end{figure}

\noindent \textbf{Effectiveness of Local-Global Iterative Editing.}
We conduct ablation studies on our LGIE training scheme in the text-driven editing task in Figure~\ref{fig:local_global}.
For the local-only setting, conducting local editing without considering the global context can lead to a discordance between the generated foreground content and the background region. 
For instance, in the second row, the right part of the sunflower is obscured by leaves. 
Conversely, performing global-only editing may change the background region, especially in complex scenes, \textit{e.g.}, the blurred background area above the dog in the first row. 
By iteratively optimizing between these two stages, we can achieve better background preservation and foreground-background harmony.

\begin{figure}[!t]
    \centering
    \includegraphics[width=\linewidth]{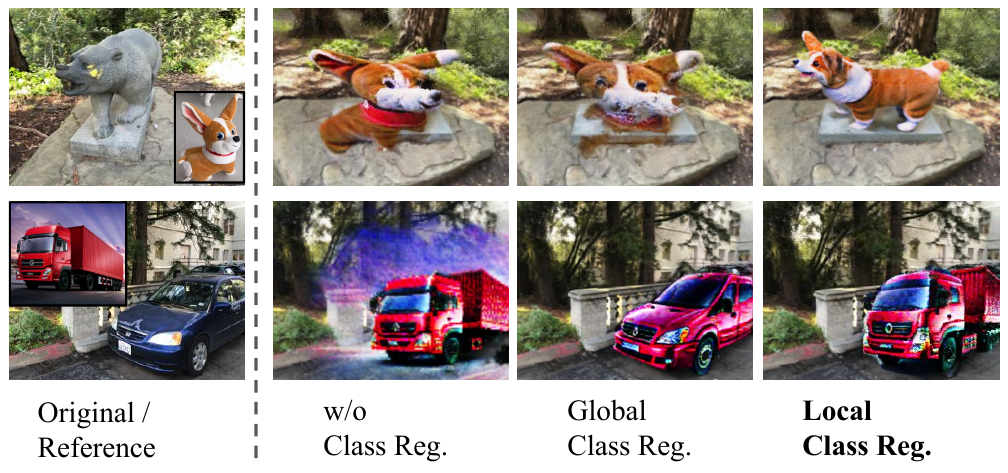}
    \caption{Ablation study of class-guided regularization. In the first column, large images depict the original scene, and small images with black borders are reference images. w/o Class Reg. denotes apply $V^*$ to both global and local stages. Global Class Reg. denotes removing $V^*$ from the global stage, and Local Class Reg. applies similar removals in the local stage.}
    \vspace{-3mm}
    \label{fig:trick}
\end{figure}

\noindent \textbf{Class-guided regularization.}
We conduct ablation studies on class-guided regularization and present the results in Figure~\ref{fig:trick}. 
Removing the class prior regularization from both local and global stages results in edited outputs that more closely resemble the reference image, but also lead to abnormal views and greater influence from the background in the reference image. 
Applying class prior to the global stage leads to an obvious Janus problem, indicating that class priors tend to work in scenarios with simpler backgrounds.
We include more ablation results in supplementary materials.

%% file: sec_arxiv/5_conclusion.tex
\section{Conclusion and Limitation}
\label{sec:conclusion}
In this study, we present the CustomNeRF model, designed for adaptive source driven 3D scene editing with unified editing prompts from either texts or reference images.
We tackle two key challenges: ensuring precise foreground-only editing, and maintaining consistency across multiple views when using single-view reference images.
Our approach includes a Local-Global Iterative Editing (LGIE) training scheme for editing that focuses on the foreground while keeping the background unchanged, and a class-guided regularization to mitigate view inconsistency in image-driven editing.
Extensive experiments validate CustomNeRF's capability to accurately edit in diverse real-world scenarios for both text and image prompts.

Our method transfers subject appearance from reference images, yet it faces limitations with Custom Diffusion's inability to perfectly replicate references, resulting in some inconsistencies in details in our image-driven editing results. Currently, our approach is also limited to text and image prompts, but future work could expand to include other editing sources like audio, sketches, etc.

%% file: sec_arxiv/appendix.tex
\clearpage
\begin{appendices}

\section{Model Architecture}
Our foreground-aware NeRF is constructed on Instant-NGP~\cite{muller2022instant} with an additional head to predict the editing probability to conduct decomposed rendering.
For simplicity in implementation, this editing probability head is designed similarly to the RGB prediction head.
It is composed of a hidden layer with 64 hidden dimensions followed by the sigmoid activation to keep the values between 0 and 1.
When rendering the foreground or background regions separately, we avoid truncating the editing probability gradient with the threshold operation.
Instead, we encourage the editing probability to approach 1 or 0 through a parameterized sigmoid activation and utilize the processed probabilities for decomposed rendering via a simple linear combination.
We find that this soft mask operation better facilitates complete shape optimization during the editing process compared to a threshold operation, as demonstrated by the ablation study presented in Section~\ref{sec:m_abla}.

\section{Implementation Details}

\noindent \textbf{Dataset preparation.}
In this paper, we conduct different editing operations using datasets such as BlendedMVS~\cite{yao2020blendedmvs} and LLFF~\cite{mildenhall2019local}. 
For 360-degree scenes from these datasets, we employ COLMAP~\cite{schoenberger2016sfm} to extract camera poses. 
This method can also be applied to real-world scenes captured by users.

\noindent \textbf{Rendering.}
Due to the memory limitation of the GPU and the high overhead of SDS loss, we need to render downsampled images.
For the BlendedMVS dataset, we render images with 3x downsampling for NeRF training and 5x downsampling for NeRF editing.
For the high-resolution LLFF and IBRNet~\cite{wang2021ibrnet} datasets, the two downsampling factors are 15 and 28.
For the bear statue dataset~\cite{in2n}, the two downsampling factors are 4 and 7.
Since we render the foreground for a separate local editing operation, we combine the rendered foreground image with a random solid color background to avoid confusion between the foreground and the single background of the same color.

\noindent \textbf{Hybrid prompt generation.} 
In image-driven editing, apart from manually inputting hybrid prompts that align with the image content, we can also employ models like GPT-4 to describe the contents of the original scene. 
To ensure that the generated descriptions adhere to our required format, we specify the output format as ``[describe the subject] in [describe the environment]'', and replace the subject part of the obtained description with ``V* [cls]''. 
Through this method, in image-driven editing, users only need to provide a reference image, eliminating the need for additional textual input to describe the original scene.

\section{More Ablation Studies}
\label{sec:m_abla}

\begin{figure}[!htbp]
    \centering
    \includegraphics[width=\linewidth]{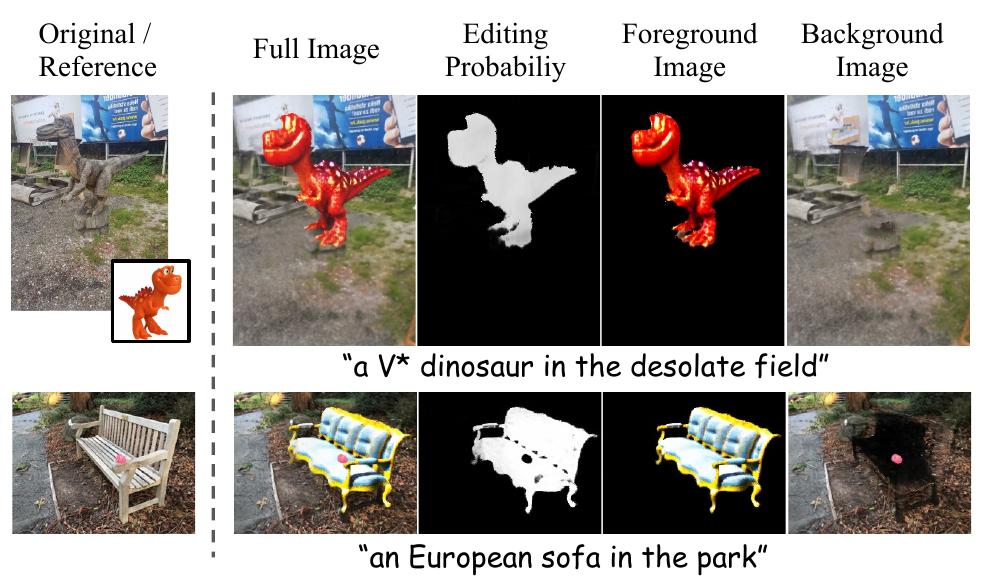}
    \caption{Visualization of intermediate results.}
    \label{fig:inter}
\end{figure}

\noindent \textbf{Visualization of intermediate results.}
We present the rendered editing probabilities and foreground/background images in Figure~\ref{fig:inter} after editing. 
It can be observed that the rendered editing probabilities can adapt well to the shape of the modified foreground regions, enabling the precise rendering of both foreground- and background-only images.
For instance, after the training process of the edited NeRF, the editing probability aligns well with the cartoon dinosaur, which is larger than the original dinosaur statue.
In the background image in the second row, we observe dark shadows in the foreground region, which indicates that NeRF does not picture the covered background regions in non-360 scenes. 
However, in the full rendering image, this shadow will be filled with foreground contents when the new objects being added are similar in shape to the original objects.

\begin{figure}[!htbp]
    \centering
    \includegraphics[width=\linewidth]{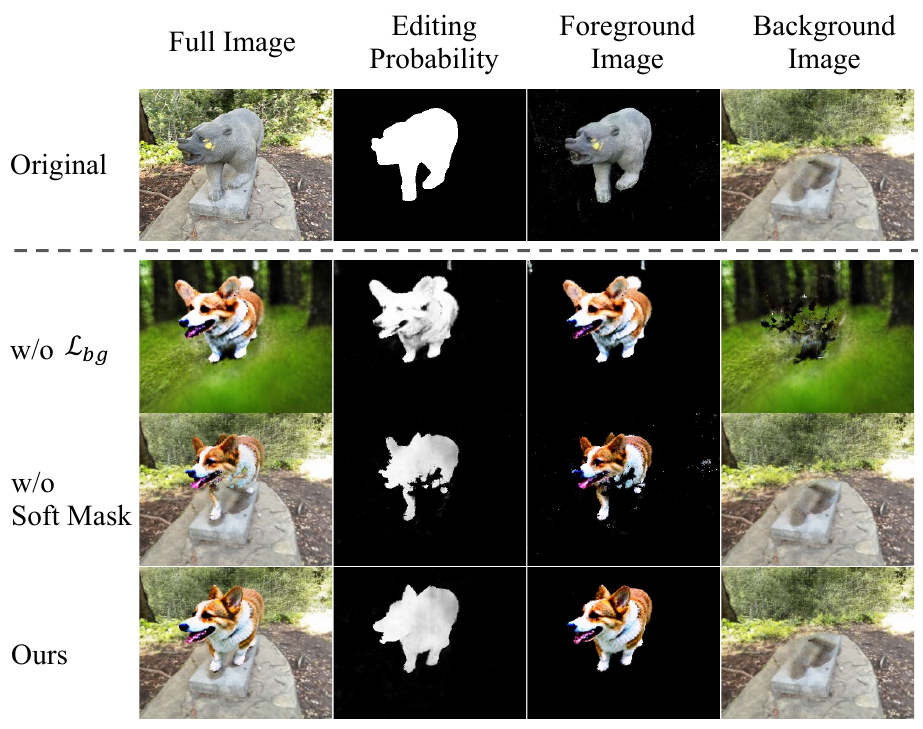}
    \caption{Ablation studies of other training strategies. w/o Soft Mask denotes utilizing a threshold to binarize the editing probabilities.}
    \label{fig:abla}
\end{figure}

\noindent \textbf{Ablation studies of other training strategies.}
We conduct more ablation studies on training strategies in Figure~\ref{fig:abla}.
Upon removal of the background preservation loss, significant changes can be observed in the background regions after editing compared with the original scene, indicating the necessity of the background preservation loss.
We also ablate different post-processing methods for editing probabilities. 
When binarizing these probabilities using a threshold (i.e., w/o Soft Mask), the obtained editing probabilities are incomplete, leading to incomplete foreground in the edited results, such as the missing left front leg of the dog in the 3-rd row of Figure~\ref{fig:abla}. 
By scaling the values between 0 and 1 using a sigmoid function to obtain a soft mask, \textit{i.e.}, the strategy adopted in our paper, we can achieve complete editing probabilities and consequently, complete editing results.

\begin{figure}[!t]
    \centering
    \includegraphics[width=\linewidth]{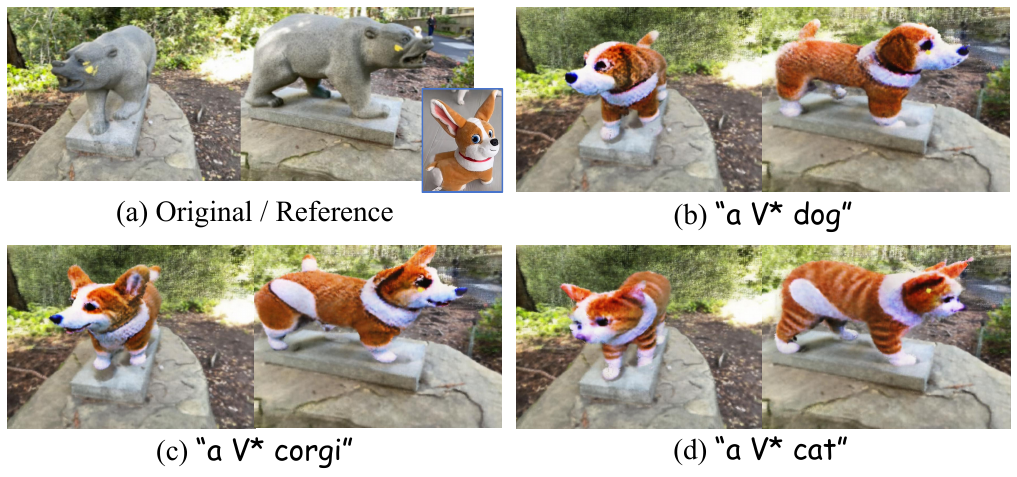}
    \caption{Visualization of different class words used in local editing prompt.}
    \label{fig:cls}
\end{figure}

\noindent \textbf{Effect of different class word.}
We use the class word in the local editing stage to promote reasonable geometry in the case of a single reference image. 
In order to explore the effect of the class word, we modify the class prior to be inconsistent with the reference image. 
As shown in Figure~\ref{fig:cls}, we compare different class words such as ``dog'', ``corgi'', and ``cat''. The final results reflect the texture of the reference image and the geometric characteristics associated with the class word we use.

\begin{figure}[!t]
    \centering
    \includegraphics[width=\linewidth]{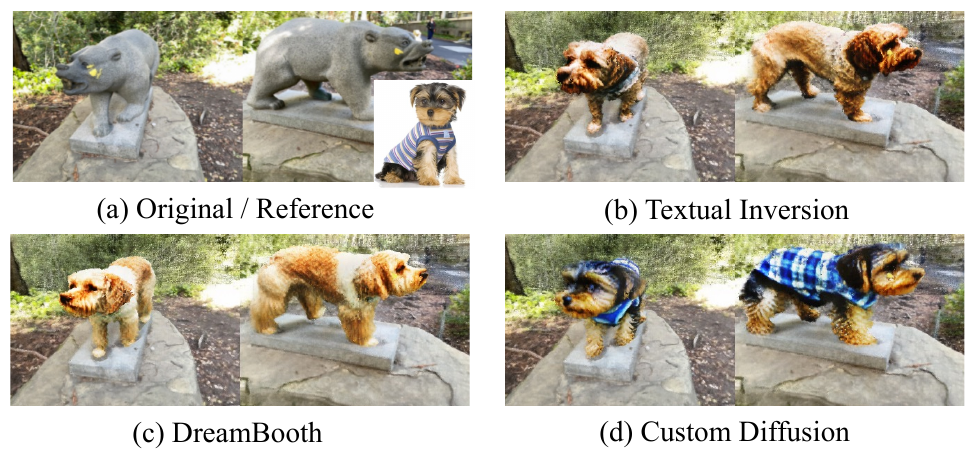}
    \caption{Comparison between different subject-aware T2I generation methods for reference subject learning.}
    \label{fig:db}
\end{figure}

\noindent \textbf{Different subject-aware T2I generation methods.}
We compare the image-driven editing results leveraging Custom Diffusion~\cite{CustomDiff} for reference subject learning with two other subject-aware T2I methods, \textit{i.e.}, Textual Inversion~\cite{TI}, and DreamBooth~\cite{dreambooth} and the visualization results are presented in Figure~\ref{fig:db}.
Given only one reference image, editing results generated with Custom Diffusion most closely resemble the reference image. 
Though the other two methods can generate similar categories and shapes, they do not match the reference image in texture. 
Therefore, in our paper, we ultimately select Custom Diffusion for reference subject learning.

\begin{figure}[!t]
    \centering
    \includegraphics[width=\linewidth]{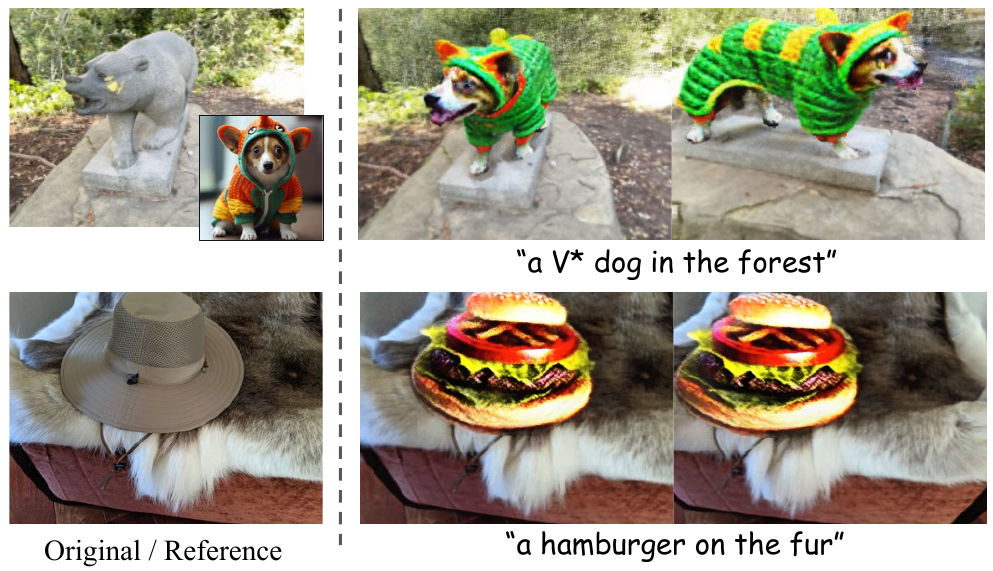}
    \caption{Failure cases.}
    \label{fig:fail}
\end{figure}

\section{More Visualization Results}

\begin{figure*}[!htbp]
    \centering
    \includegraphics[width=.95\linewidth]{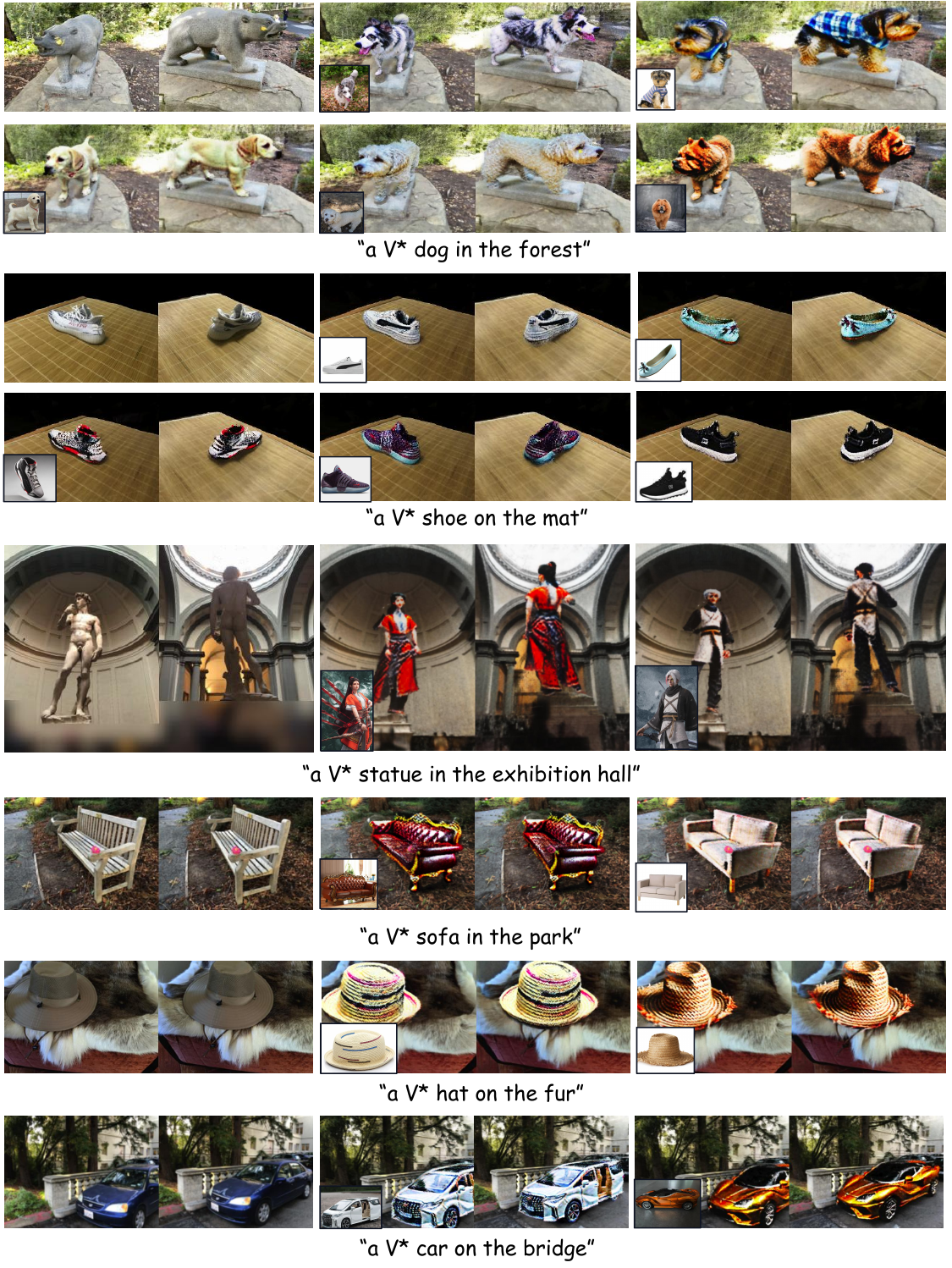}
    \caption{More visualization results of image-driven editing.}
    \label{fig:m_img}
\end{figure*}

\begin{figure*}[!htbp]
    \centering
    \includegraphics[width=.95\linewidth]{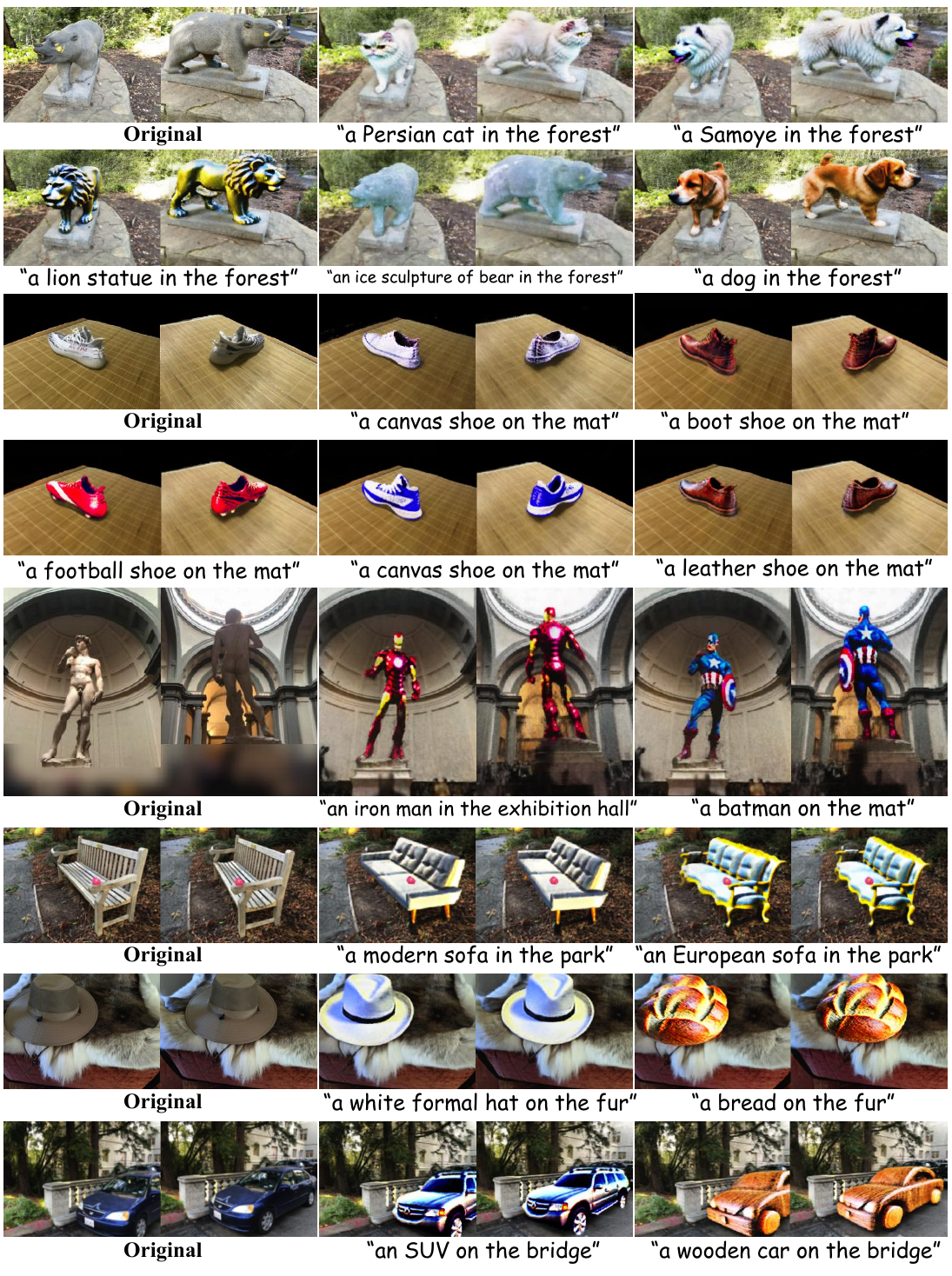}
    \caption{More visualization results of text-driven editing.}
    \label{fig:m_text}
\end{figure*}

\noindent \textbf{Qualitative results.}
We provide more image-driven and text-driven editing results in Figure~\ref{fig:m_img} and Figure~\ref{fig:m_text} respectively.
We have also included video editing results in the ``video results'' folder, which provides dynamic visualization of editing results.

\noindent \textbf{Failure cases.}
In addition to the experimental results above, we also present the failure cases of CustomNeRF in Figure~\ref{fig:fail}. 
For instance, in the first row, as the Custom Diffusion cannot always generate images with the subject identical to those in the reference image, the edited scene shows the dog's clothing different from that in the reference image.
In the second row, due to the considerable shape difference between a hat and a hamburger, the replaced hamburger takes on an inverted shape, larger at the bottom and smaller at the top, similar to the hat's structure.

\end{appendices}